\def\year{2022}\relax
\documentclass[letterpaper]{article} 
\usepackage{aaai22}  
\usepackage{times}  
\usepackage{helvet}  
\usepackage{courier}  
\usepackage[hyphens]{url}  
\usepackage{graphicx} 
\def\UrlFont{\rm}  
\usepackage{natbib}  
\usepackage{caption} 
\DeclareCaptionStyle{ruled}{labelfont=normalfont,labelsep=colon,strut=off} 
\usepackage{algorithm}
\usepackage{algorithmic}

\usepackage{newfloat}
\usepackage{listings}
\floatstyle{ruled}
\newfloat{listing}{tb}{lst}{}
\floatname{listing}{Listing}

\usepackage{xcolor}

\usepackage{amsmath,amsfonts,bm}

\def\eqref#1{equation~\ref{#1}}

\def\1{\bm{1}}

\DeclareMathAlphabet{\mathsfit}{\encodingdefault}{\sfdefault}{m}{sl}
\SetMathAlphabet{\mathsfit}{bold}{\encodingdefault}{\sfdefault}{bx}{n}

\newcommand{\Ls}{\mathcal{L}}

\newcommand{\KL}{D_{\mathrm{KL}}}

\newcommand{\ie}{\textit{i}.\textit{e}.}
\newcommand{\eg}{\textit{e}.\textit{g}.}

\newcommand{\tempcite}[1]{\nocite{#1}}

\newcommand{\Ps}{p^s}
\newcommand{\Pt}{p^t}
\newcommand{\Zs}{z^s}
\newcommand{\Zt}{z^t}

\iftrue
    \newcommand{\zk}[1]{}
    \newcommand{\yh}[1]{}
    \newcommand{\ys}[1]{}
    \newcommand{\js}[1]{}
    \newcommand{\revised}[1]{#1}
\else
    \newcommand{\zk}[1]{{\color{blue}{(Zsolt: #1)}}}
    \newcommand{\yh}[1]{{\color{red}{(Yen: #1)}}}
    \newcommand{\ys}[1]{{\color{cyan}{(Yilin: #1)}}}
    \newcommand{\js}[1]{{\color{green}{(James: #1)}}}
    \newcommand{\revised}[1]{{\color{orange}{#1}}}
\fi

\usepackage{booktabs}
\usepackage{multirow}
\usepackage{subfig}
\usepackage{rotating}

\title{\revised{A Closer Look at Knowledge Distillation with Features, Logits, and Gradients}}
\author{
    Yen-Chang Hsu\textsuperscript{\rm 1}, James Smith\textsuperscript{\rm 2}, Yilin Shen\textsuperscript{\rm 1},  Zsolt Kira\textsuperscript{\rm 2}, Hongxia Jin\textsuperscript{\rm 1}
}
\affiliations{
    \textsuperscript{\rm 1}Samsung Research America,
    \textsuperscript{\rm 2}Georgia Institute of Technology\\
    \{yenchang.hsu, yilin.shen, hongxia.jin\}@samsung.com, \{jamessealesmith, zkira\}@gatech.edu
}

\begin{document}

\maketitle

\begin{abstract}
Knowledge distillation (KD) is a substantial strategy for transferring learned knowledge from one neural network model to another. A vast number of methods have been developed for this strategy.
While most method designs a more efficient way to facilitate knowledge transfer, less attention has been put on comparing the effect of knowledge sources such as features, logits, and gradients. 
\revised{This work provides a new perspective to motivate a set of knowledge distillation strategies by approximating the classical KL-divergence criteria with different knowledge sources, making a systematic comparison possible in model compression and incremental learning. 
Our analysis indicates that logits are generally a more efficient knowledge source and suggests that having sufficient feature dimensions is crucial for the model design, providing a practical guideline for effective KD-based transfer learning.}
\end{abstract}

\section{Introduction}
Knowledge distillation transfers knowledge from one neural network model to another by matching different information sources from the models: logits \cite{hinton2014distilling}, features \cite{romero2014fitnets}, or gradients \cite{srinivas2018knowledge}. It has found wide applications in such areas as model compression \cite{hinton2014distilling, wang2020knowledge, ba2014deep, urban2016deep}, incremental learning \cite{li2016learning, lee2019overcoming}, privileged learning \cite{lopez2015unifying}, adversarial defense \cite{papernot2016distillation},
and learning with noisy data \cite{li2017learning}. This knowledge transfer usually follows the teacher-student scheme, in that a high-performing teacher model provides knowledge sources for a student model to match.

\revised{Which part of the teacher provides a more informative knowledge source for a student to distill? \citet{heo2019knowledge} shows the features are more effective; \citet{tian2019crd} observes that logits are generally better, but matching the pairwise correlation with features outperforms logits. Recently, \citet{kim2021comparing} reports that logits achieve a better result when using L2 loss instead of KL-divergence. This puzzling inconsistency is due to the differences of both optimization criteria and the knowledge sources used in different methods, making a fair comparison across knowledge sources impossible.

This work provides a systematic approach to analyzing the above puzzle, providing a closer look at transferring knowledge in features, logits, and gradients. Specifically, we propose a new perspective to reinterpret the classical KD objective, using Taylor expansion to approximate the KL-divergence with different knowledge sources.
This novel perspective leads to a generalized divergence that allows us to use logits and features under the same optimization criteria, while the gradients provide information about the \textit{importance} of each feature for distillation.
We therefore based on the mathematical insight as a unified framework, instantiating varied methods to distill features, logits, and gradients. Interestingly, when we instantiate our variants with the simplest design choices, the resulting methods are similar to previous knowledge transfer techniques. The similarity demonstrates the generalizability of our framework. Nevertheless, there are still nuanced differences between our variants and prior methods to ensure a fair comparison across knowledge sources, discussed later.

We explore two aspects that have never been jointly investigated with knowledge sources. The first is observing whether the trend of effectiveness is consistent across different transfer learning tasks. The two most popular KD tasks are included: model compression and incremental learning. The second aspect is to investigate the impact of model design and identify the key factors affecting knowledge sources' effectiveness. In summary, this work provides the following contributions and findings:
}
\begin{itemize}
    \item \revised{We provide a new perspective to interpret the classical KD, showing that a single framework can describe the distillation with different knowledge sources. To the best of our knowledge, this perspective has not been presented.}
    \item \revised{The new perspective leads to a new strategy for improving feature-based KD by weighing the importance of features based on gradients from the teacher.}
    \item Our systematic comparison shows that logits generally is the more effective knowledge source, followed by the features weighted by gradients and the plain features. This trend is consistent in both model compression and incremental learning.
    \item We further use a new metric with a normalized basis to analyze the factors that affect the above trend, pointing out that the feature size of the penultimate layer plays a crucial role in impacting different knowledge sources' effectiveness.
\end{itemize}

\section{Rethinking Knowledge Distillation} \label{sec:rethinking}

Our observation starts with the generic knowledge distillation criteria $\Ls_{KD}$ for classification \cite{hinton2014distilling}. The criteria includes cross-entropy loss $\Ls_{CE}$ for a student model to learn from ground-truth label $y^*$, and $\KL$ for minimizing the difference between the predicted class distribution $\Ps$ (from the student) and $\Pt$ (from the teacher). The latter term makes the knowledge transfer happen, for which the coefficient $\lambda$ controls the intensity:
\begin{equation} \label{eq:KD}
\Ls_{KD}=\Ls_{CE}(\Ps, y^*) + \lambda\KL(\Pt||\Ps),
\end{equation}
\begin{equation} \label{eq:KLdiv}
\KL(\Pt || \Ps)=\sum_y \Pt_y \log \Pt_y - \sum_y \Pt_y \log \Ps_y
\end{equation}

The next step is to expand $\KL$. Here, we are more interested in how intermediate outputs $z$ from the teacher ($\Zt=g^t(x)$) and student ($\Zs=g^s(x)$) affect $\KL$; therefore, we treat the $z$ as the only variable in $\Pt_y=f(y;\Zt)$ and $\Ps_y=f(y;\Zs)$, leaving the parameters (if any) of the softmax-based classifier $f$ as constants and the same $f$ is used for both the teacher and student.
By taking the Taylor expansion around $\Zt$ for the second term of \eqref{eq:KLdiv} and using the notation $dz=\Zs-\Zt$, $\KL$ becomes:
\begin{align} 
\KL(\Pt || \Ps)=&\sum\Pt\log\Pt-\sum\Pt\log\Pt\nonumber\\
&-dz^T\sum \Pt \frac{d}{dz}\log \Pt\nonumber\\
&-\frac{1}{2}dz^T(\sum\Pt\frac{d^2}{dz^2}\log\Pt) dz+\epsilon\label{eq:KLrawexpand}\\
=&\frac{1}{2}dz^T\sum\Pt(\frac{d}{dz}\log\Pt)(\frac{d}{dz}\log\Pt)^T dz+\epsilon \label{eq:KLexpand}
\end{align}
In \eqref{eq:KLrawexpand}, the first-order term is zero, while the second-order has a form of Fisher information matrix $F(\Zt)$ at its middle. Details of derivation are available in the Supplementary. The above equations omit the $y$ for conciseness. Lastly, by ignoring the higher-order term $\epsilon$, $\KL$ can be rewritten as:
\begin{equation} \label{eq:KLapprox}
\KL(\Pt || \Ps)\approx \frac{1}{2}(\Zs-\Zt)^T F(\Zt) (\Zs-\Zt)
\end{equation}
Although \eqref{eq:KLapprox} has a simple quadratic form, it provides two key insights:
\begin{enumerate}
  \item Minimizing the difference between the student's and teacher's intermediate representations reduces the KL-divergence.
  \item The Fisher information $F(\Zt)$, which leverages the gradients regarding the teacher's intermediate representation, provides a weighting mechanism for the importance of features. 
\end{enumerate}

\section{The Generalized Divergence} \label{sec:framework}

\revised{This section discusses the simplest design choices to turn \eqref{eq:KLapprox} into a framework that is easy to implement and instantiate its variants.}
There are two empirical considerations in applying \eqref{eq:KLapprox}, both stemming from the challenges of computing $F(\Zt)$. The first is its $O(|z|^2)$ complexity. The computation could be expensive when $z$ has a large dimension (\eg the flattened feature map from a convolution neural network based on image inputs). Here we follow the common simplification used by EWC~\cite{kirkpatrick2017overcoming} and Adam~\cite{kingma2014adam} in which only the diagonal of Fisher information matrix is considered, reducing the complexity to $O(|z|)$. Second, marginalizing over $y$ for accumulating the gradients could be time-consuming; therefore, there is a need to use an alternative loss function to collect gradients of $\Zt$. 

To adopt the above considerations, we define a generalized divergence $D_G$, making \eqref{eq:KLapprox} a special case of the generalized form:
\begin{equation} \label{eq:D_G}
D_G(\Zt, \Zs)=\alpha(\Zs-\Zt)^T W(\Zt) (\Zs-\Zt)
\end{equation}
The coefficient $\alpha$ is a scaling factor that can be absorbed by $\lambda$. The $W(\Zt)$ is still an n-by-n weighting matrix like $F(\Zt)$, given $\Zt$'s dimension $n$. The calculation of $W$ is still based on gradients from the teacher, specifically:
\begin{equation}
W(\Zt)=diag((\frac{d}{dz}\Ls_*)(\frac{d}{dz}\Ls_*)^T)\label{eq:W}    
\end{equation}
The function $diag$ casts all off-diagonal elements to be zero. The $\Ls_*$ is $\log\Pt$ when computing the Fisher information. Here we consider two design choices for $\Ls_*$ to avoid the need of marginalizing over $y$:
\begin{align}
\Ls_E&=\log\Pt_{y^*}\\
\Ls_H&=\frac{1}{k}\sum_{y=1}^k (l^t_y)^2
\end{align}
$\Ls_E$ is related to the empirical Fisher which requires knowing the ground-truth class $y^*$. $\Ls_H$ is a heuristic criteria by using the mean-squared logits ($l_y$) over $k$ classes. $\Ls_H$ does not require labels,
but captures the gradients that lead to a large change in logits. $\Ls_H$ is useful when the student (and its training data) has a different set of classes from the teacher, which is a case that $\Ls_E$ is not applicable.

Overall, our full criteria $\Ls_{KD-G}$ with the generalized divergence has the form:
\begin{equation} \label{eq:ourKD}
\Ls_{KD-G}=\Ls_{CE}(\Ps, y^*) + \lambda D_G(\Zt, \Zs)
\end{equation}
Note $z$ can be the logits or features. Besides, the knowledge in the teacher's gradients are transferred to the student via $W(\Zt)$. Therefore $\Ls_{KD-G}$ provides a \textit{unified} framework for comparing the effectiveness of each knowledge source by instantiating it in different ways. This is one of the main contributions of this paper, as our formulation allows an explicit fair comparison across knowledge sources within a unified framework. Below we elaborate the cases when $z$ is features or logits, and additionally extend the discussion to a model's parameters, which is a popular knowledge source in the incremental learning.

\begin{figure}
  \centering
  \includegraphics[clip, trim=0cm 0.4cm 16cm 0cm, width=0.5\textwidth]{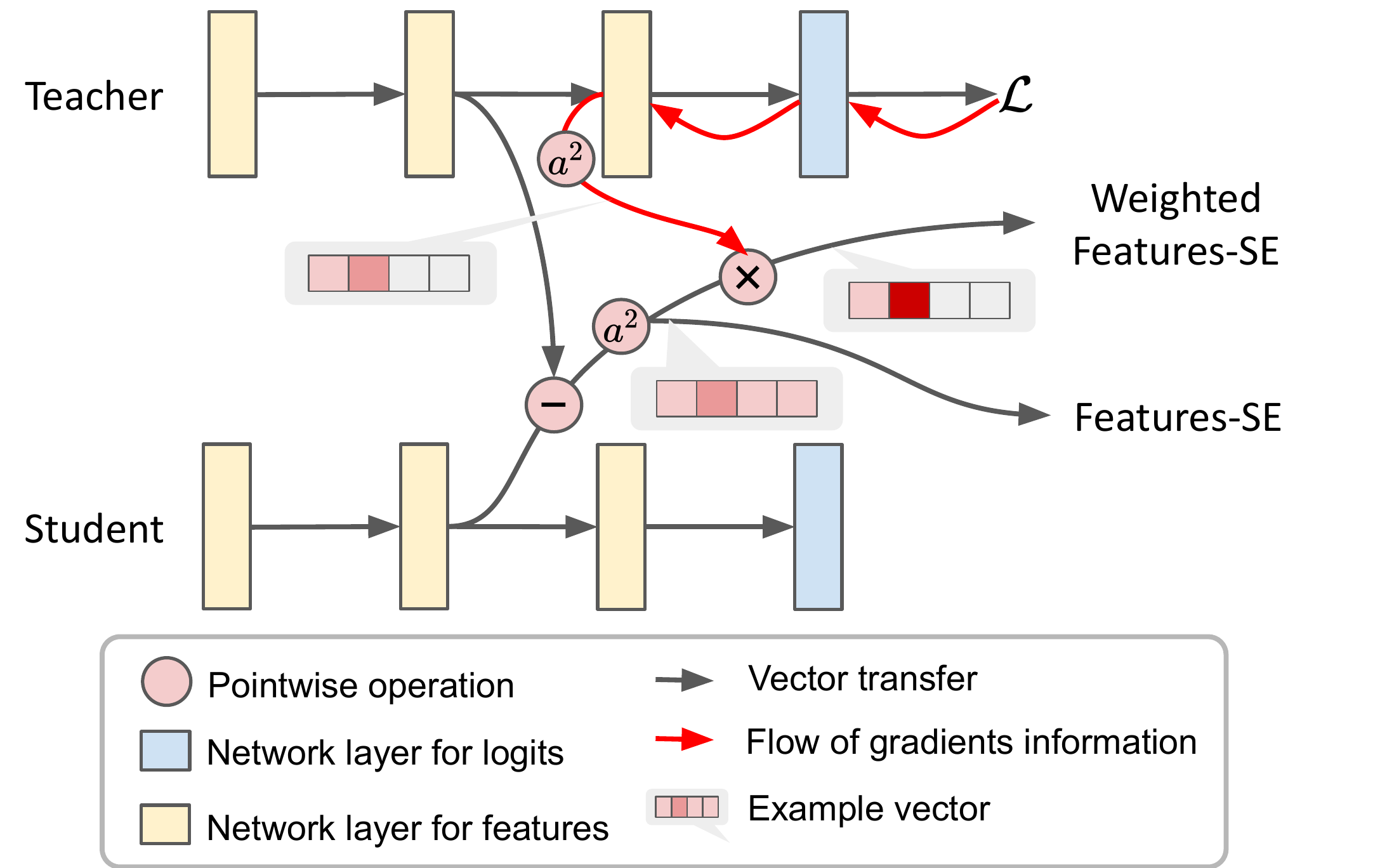} 
  \caption{The schematic illustration for the variants that use features. The example vectors demonstrate how the gradients weigh the features. A darker color in the example vector indicates the feature is more influential. Note that the operation $a^2$ means all elements in the vector are squared. SE is the abbreviation of squared error.
  }\label{fig:framework}
\end{figure}

\subsection{Features}\label{sec:features}

When $z$ is features, there are rich choices for computing $W$. One can use $\Ls_E$ when the ground-truth class $y^*$ is available. Such a condition is usually true in the case of model compression. In contrast, there may not be a valid label for the teacher model (as we will see in the later section on incremental learning).
In such cases, we use $\Ls_H$ for $W$. Furthermore, $W$ can also be an identity matrix, which reduces $D_G$ to a simple squared error. This case leads to the same optimization objective used in FitNet \cite{romero2014fitnets}. Therefore, we include $W=\mathbb{I}$ as one of the variants in our framework while simplifying its deployment with a principled normalization to make it amicable to KD tasks other than model compression, described in the later sections.

\subsection{Logits}\label{sec:logits}
When $z$ is logits, its Hessian matrix (as appears in \eqref{eq:KLrawexpand}) obtained through $\Ls_H$ is an identity matrix ($W=\mathbb{I}$ with a scaling factor that can be absorbed by the coefficient $\lambda$), indicating that a weighting mechanism based on gradients is redundant for this case. As a result, $D_G$ reduces to a simple squared error with logits. The result has the same formula as when computing the KL-divergence between the output probabilities with the logits being divided by a large temperature \cite{hinton2014distilling, kim2021comparing}, indicating this simple variant is worth more attention in a comparative study.

\subsection{Parameters} \label{sec:param_reg}
In an alternative perspective, which switches the focus from intermediate representations $z$ to model's parameters $\theta^t$ (from $g^t$) and $\theta^s$ (from $g^s$), a derivation similar to \eqref{eq:KLrawexpand} will lead to the parameter-based regularization. Note that our process derives it from a different motivation, but reaches the same formulation of EWC~\cite{kirkpatrick2017overcoming}, an importance-weighted parameter regularization method:
\begin{equation} \label{eq:EWC}
\Ls_{EWC}=\Ls_{CE}(\Ps, y^*) + \frac{\lambda}{2}(\theta^s-\theta^t)^T F(\theta^t) (\theta^s-\theta^t)
\end{equation}

Although $\Ls_{KD-G}$ and $\Ls_{EWC}$ share a similar form, they have four fundamental differences:
(1) $\Ls_{KD-G}$ is for features or logits, while $\Ls_{EWC}$ is only for parameters.
(2) Our derivation starts from KL-divergence between categorical distributions, while EWC starts from the normal approximation of the posterior for parameters.
(3) EWC requires the teacher and student model to be the same. As a result, EWC is not considered as a KD method and is not applicable to model compression. (4) $\Ls_{KD-G}$ achieves significantly better results than $\Ls_{EWC}$ in the task-incremental learning (shown in the experiment section).

\subsection{The Four Instantiations
} \label{sec:variants}
Based on the above discussion, we create four $\Ls_{KD-G}$ instantiations to enable a systematic comparison. The first two (Weighted$_E$ Features-SE and Weighted$_H$ Features-SE) are novel KD methods derived from our framework, while the latter two (Features-SE and Logits-SE) have close alternatives in previous works. The illustration of the first three variants is shown in Figure \ref{fig:framework}. Their names are listed below with a description of how their $W$ is implemented for the $\Ls_{KD-G}$ (Note: SE means squared error):
\begin{itemize}
  \item Weighted$_E$ Features-SE: $W$ uses $\Ls_E$
  \item Weighted$_H$ Features-SE: $W$ uses $\Ls_H$
  \item Features-SE: $W=\mathbb{I}$
  \item Logits-SE: $W=\mathbb{I}$
\end{itemize}

\section{Experiments}

\subsection{Model Compression (MC)} \label{sec:MC}

Model compression is the primary task where knowledge distillation techniques are applied heavily. In this case, the student model has a smaller capacity than the teacher yet is asked to match the teacher's outputs. If the target for matching contains a great deal of  information that is not crucial for a downstream task, the student is more likely to waste its limited capacity on matching unimportant information, capping the student's ability to reach the teacher's performance. This argument gives an intuition of why one can expect the weighted features to perform better than an unweighted one. A similar argument may apply to the logits since the linear layer before the logits imposes weights on the features. This section provides empirical support for the arguments with our unified framework.

\subsubsection{Applying our framework}
Two empirical considerations need to be addressed for applying \eqref{eq:ourKD} to model compression. The first one is that a large numerical range can result from $D_G$, potentially making the optimization unstable. In the KD training procedure, the student is initialized randomly and is directly optimized from scratch with the knowledge distillation criteria. The random student model could make the squared difference between $\Zs$ and $\Zt$ unbounded. This issue can be addressed by normalization. Specifically, we make $\hat{z}^s$ and $\hat{z}^t$ unit vectors:
\begin{equation} \label{eq:norm_Z}
\hat{z}^t=\frac{\Zt}{||\Zt||}, \hat{z}^s=\frac{\Zs}{||\Zs||}
\end{equation}

The second empirical consideration is the mismatched dimensions between $\Zs$ and $\Zt$ when they are features. This case happens when the student and teacher have different types of neural network architectures or when the student has a smaller model width. We add a linear transformation $r$ on the outputs of  $g^s(x)$ to match the teacher's dimension:
\begin{equation} \label{eq:transform_z}
\Zs=r(g^s(x))
\end{equation}
The parameters of $r$ are also optimized by the customized $D_G$ for this section:
\begin{equation} \label{eq:KD_compression}
D_G^{MC}=(\hat{z}^s-\hat{z}^t)^T W(\Zt) (\hat{z}^s-\hat{z}^t)
\end{equation}
Note that $r$ is only involved during training and is removed from testing; thus, the student model's design has no dependency on $r$. Additionally, $r$ is only used when z is features, since the logits layer always has the same dimensionality (number of classes) between the teacher and student.

\begin{table*}
\begin{center}
\begin{small}
\begin{tabular}{lcccccccc}
\toprule
Teacher model         &        & resnet32x4       & vgg13            & WRN-40-2         & WRN-40-2         & resnet110          \\
Student model         &.       & resnet8x4        & vgg8             & WRN-16-2         & WRN-40-1         & resnet32           \\
\midrule
Teacher acc.          &        & 79.42            & 74.64            & 75.61            & 75.61            & 74.31              \\
Student acc.          &        & 72.50            & 70.36            & 73.26            & 71.98            & 71.14              \\
\toprule
Method               & Knowledge source & & & & & \\
\midrule
HKD (Hinton 2014) \tempcite{hinton2014distilling}                    & L      & 73.33 $\pm$ 0.25 & 72.98 $\pm$ 0.19 & 74.92 $\pm$ 0.28 & 73.54 $\pm$ 0.20 & 73.08 $\pm$ 0.18   \\
FitNet (Romero 2015)\tempcite{romero2014fitnets}                & F      & 73.50 $\pm$ 0.28 & 71.02 $\pm$ 0.31 & 73.58 $\pm$ 0.32 & 72.24 $\pm$ 0.24 & 71.06 $\pm$ 0.13   \\
AT (Zagoruyko 2017)\tempcite{zagoruyko2016paying}                   & F      & 73.44 $\pm$ 0.19 & 71.43 $\pm$ 0.09 & 74.08 $\pm$ 0.25 & 72.77 $\pm$ 0.10 & 72.31 $\pm$ 0.08   \\
FT (Kim 2018)\tempcite{kim2018paraphrasing}                   & F      & 72.86 $\pm$ 0.12 & 70.58 $\pm$ 0.08 & 73.25 $\pm$ 0.20 & 71.59 $\pm$ 0.15 & 72.37 $\pm$ 0.31   \\
SP (Tung 2019)\tempcite{tung2019similarity}                   & F$^+$  & 72.94 $\pm$ 0.23 & 72.68 $\pm$ 0.19 & 73.83 $\pm$ 0.12 & 72.43 $\pm$ 0.27 & 72.69 $\pm$ 0.41   \\
CRD (Tian 2020)\tempcite{tian2019crd}                  & F$^+$  & 75.51 $\pm$ 0.18 & 73.94 $\pm$ 0.22 & 75.48 $\pm$ 0.09 & 74.14 $\pm$ 0.22 & 73.48 $\pm$ 0.13   \\
CRD+HKD \tempcite{tian2019crd}               &L + F$^+$& 75.46 $\pm$ 0.25 & 74.29 $\pm$ 0.12 & 75.64 $\pm$ 0.21 & \textbf{74.38} $\pm$ 0.11 & \textbf{73.75} $\pm$ 0.24   \\
\midrule
(A) Features-SE       & F      & 74.83 $\pm$ 0.15 & 73.08 $\pm$ 0.26 & 75.35 $\pm$ 0.20 & 74.06 $\pm$ 0.43 & 73.13 $\pm$ 0.18   \\	
(B) Weighted$_E$ Features-SE & F + G  & 75.20 $\pm$ 0.15 & 73.57 $\pm$ 0.22 & 75.33 $\pm$ 0.30 & 73.87 $\pm$ 0.39 & 73.37 $\pm$ 0.26   \\
(C) Logits-SE         & L      & 76.29 $\pm$ 0.16 & \textbf{74.45} $\pm$ 0.19 & 75.67 $\pm$ 0.25 & \textbf{74.38} $\pm$ 0.16 & 73.21 $\pm$ 0.31   \\
(B+C)                 &L + F + G& \textbf{76.66} $\pm$ 0.22 & 74.28 $\pm$ 0.14 & \textbf{75.74} $\pm$ 0.20 & 74.29 $\pm$ 0.11 & 73.48 $\pm$ 0.24   \\
\bottomrule						
\end{tabular}
\end{small}
\caption{
Test accuracy (\%) of student models on CIFAR100 of many distillation methods. The teacher-student pair is from the \textbf{same} architecture family. The variants from our framework are labeled with (A), (B), (C), and (B+C). The results of other methods are inherited from CRD (a state-of-the-art method) since we strictly follow its training and testing procedure and can reproduce its reported results. The type of knowledge sources is categorized by: logits (L), features (F), higher-order features (\ie, instance-wise correlation, F$^+$), and gradients (G). 
}
\label{tbl:cifar100_same}
\end{center}
\end{table*}
\begin{table*}[hbt!]

\begin{center}

\begin{small}
\begin{tabular}{lccccccc}
\toprule
Teacher model         &             & resnet32x4       & resnet32x4       & WRN-40-2         & ResNet50         & ResNet50         & vgg13             \\
Student model         &             & ShuffleNetV1     & ShuffleNetV2     & ShuffleNetV1     & vgg8             & MobileNetV2      & MobileNetV2       \\
\midrule             
Teacher acc.          &             & 79.42            & 79.42            & 75.61            & 79.34            & 79.34            & 74.64            \\
Student acc.          &             & 70.5             & 71.82            & 70.5             & 70.36            & 64.6             & 64.6              \\
\toprule     
Method                & Source       & & & & & \\
\midrule           
HKD (Hinton 2014) \tempcite{hinton2014distilling}                   & L           & 74.07 $\pm$ 0.19 & 74.45 $\pm$ 0.27 & 74.83 $\pm$ 0.17 & 73.81 $\pm$ 0.13 & 67.35 $\pm$ 0.32 & 67.37 $\pm$ 0.32  \\
FitNet (Romero 2015) \tempcite{romero2014fitnets}                & F           & 73.59 $\pm$ 0.15 & 73.54 $\pm$ 0.22 & 73.73 $\pm$ 0.32 & 70.69 $\pm$ 0.22 & 63.16 $\pm$ 0.47 & 64.14 $\pm$ 0.50  \\
AT (Zagoruyko 2017)\tempcite{zagoruyko2016paying}                    & F           & 71.73 $\pm$ 0.31 & 72.73 $\pm$ 0.09 & 73.32 $\pm$ 0.35 & 71.84 $\pm$ 0.28 & 58.58 $\pm$ 0.54 & 59.40 $\pm$ 0.20  \\
FT (Kim 2018)\tempcite{kim2018paraphrasing}                    & F           & 71.75 $\pm$ 0.20 & 72.50 $\pm$ 0.15 & 72.03 $\pm$ 0.16 & 70.29 $\pm$ 0.19 & 60.99 $\pm$ 0.37 & 61.78 $\pm$ 0.33  \\
SP (Tung 2019)\tempcite{tung2019similarity}                    & F$^+$       & 73.48 $\pm$ 0.42 & 74.56 $\pm$ 0.22 & 74.52 $\pm$ 0.24 & 73.34 $\pm$ 0.34 & 68.08 $\pm$ 0.38 & 66.30 $\pm$ 0.38  \\
CRD (Tian 2020)\tempcite{tian2019crd}                  & F$^+$       & 75.11 $\pm$ 0.32 & 75.65 $\pm$ 0.10 & 76.05 $\pm$ 0.14 & 74.30 $\pm$ 0.14 & 69.11 $\pm$ 0.28 & 69.73 $\pm$ 0.42  \\
CRD+HKD \tempcite{tian2019crd}                &L + F$^+$    & 75.12 $\pm$ 0.35 & 76.05 $\pm$ 0.09 & 76.27 $\pm$ 0.29 & 74.58 $\pm$ 0.27 & 69.54 $\pm$ 0.39 & \textbf{69.94} $\pm$ 0.05  \\
\midrule             
(A) Features-SE       & F           & 75.47 $\pm$ 0.13 & 75.61 $\pm$ 0.25 & 75.91 $\pm$ 0.33 & 73.78 $\pm$ 0.31 & 68.38 $\pm$ 0.80 & 68.24 $\pm$ 0.13  \\
(B) W$_E$ Features-SE & F + G       & 76.24 $\pm$ 0.29 & 76.14 $\pm$ 0.23 & 76.30 $\pm$ 0.15 & 74.42 $\pm$ 0.19 & 69.29 $\pm$ 0.21 & 68.36 $\pm$ 0.30  \\
(C) Logits-SE         & L           & 75.90 $\pm$ 0.47 & \textbf{76.64} $\pm$ 0.18 & 76.36 $\pm$ 0.24 & 75.02 $\pm$ 0.35 & 69.95 $\pm$ 0.18 & 68.99 $\pm$ 0.52  \\
(B+C)                 &L + F + G    & \textbf{76.99} $\pm$ 0.43 & 76.56 $\pm$ 0.35 & \textbf{76.91} $\pm$ 0.29 & \textbf{75.08} $\pm$ 0.28 & \textbf{70.58} $\pm$ 0.26 & 69.79 $\pm$ 0.33  \\
\bottomrule
\end{tabular}
\end{small}
\caption{
Test accuracy (\%) of student models on CIFAR100 with transfer across very \textbf{different} teacher and student architectures.
}
\label{tbl:cifar100_diff}
\end{center}

\end{table*}
\begin{table*}
\begin{center}
\begin{small}
\begin{tabular}{lccccc}
\toprule
Teacher model            &                  & \multicolumn{2}{c}{ResNet50}                          & \multicolumn{2}{c}{ResNet34}                          \\
Student model            &                  & \multicolumn{2}{c}{ResNet18}                          & \multicolumn{2}{c}{ResNet18}                          \\
Method                   & Knowledge source & \multicolumn{1}{c}{Top 1} & \multicolumn{1}{c}{Top 5} & \multicolumn{1}{c}{Top 1} & \multicolumn{1}{c}{Top 5} \\
\toprule
Teacher                 & -                & 62.22                     & 83.52                     & 57.86                     & 80.39                     \\
Student (baseline)                 & -                & 53.15                     & 75.97                     & 53.15                     & 75.97                     \\
\midrule
HKD (Hinton 2014) \tempcite{hinton2014distilling}                     & L                & 58.26 $\pm$ 0.34                   & 80.92 $\pm$ 0.25                   & 58.54 $\pm$ 0.24                   & 81.07 $\pm$ 0.12                   \\
AT (Zagoruyko 2017) \tempcite{zagoruyko2016paying}                      & F                & 55.67 $\pm$ 0.32                   & 79.40 $\pm$ 0.22                   & 56.04 $\pm$ 0.24                   & 79.09 $\pm$ 0.35                   \\
SP (Tung 2019) \tempcite{tung2019similarity}                      & F$^+$               & 55.48 $\pm$ 0.33                   & 78.34 $\pm$ 0.31                   & 55.28 $\pm$ 0.32                   & 78.46 $\pm$ 0.29                   \\
PKT (Passalis 2018) \tempcite{pkt_eccv}                     & F$^+$               & 54.34 $\pm$ 0.20                   & 77.49 $\pm$ 0.24                   & 54.45 $\pm$ 0.42                   & 77.48 $\pm$ 0.21                   \\
RKD (Park 2019) \tempcite{park2019relational}                     & F$^+$               & 54.30 $\pm$ 0.44                   & 77.31 $\pm$ 0.43                   & 54.37 $\pm$ 0.22                   & 77.40 $\pm$ 0.30                   \\
\midrule
(A) Features-SE          & F                & 56.86 $\pm$ 0.21                   & 79.44 $\pm$ 0.38                   & 56.57 $\pm$ 0.40                   & 79.43 $\pm$ 0.22                   \\
(B) Weighted$_E$ Features-SE & F + G            & 57.15 $\pm$ 0.29                   & 79.55 $\pm$ 0.24                   & 57.08 $\pm$ 0.17                   & 79.54 $\pm$ 0.36                   \\
(C) Logits-SE            & L                & 58.95 $\pm$ 0.34                   & 81.14 $\pm$ 0.12                   & 58.53 $\pm$ 0.32                   & 80.80 $\pm$ 0.31                   \\
(B+C)                    & L + F + G        & \textbf{59.30} $\pm$ 0.20                   & \textbf{81.46} $\pm$ 0.15                   & \textbf{58.79} $\pm$ 0.15                   & \textbf{81.09} $\pm$ 0.06 \\                 
\bottomrule
\end{tabular}
\end{small}
\caption{
Test accuracy (\%) of student models on \textbf{Tiny-ImageNet}. The type of knowledge sources is categorized by: logits (L), features (F), higher-order features (\ie, instance-wise correlation, F$^+$), and gradients (G). The value is averaged with 5 repeats. 
}\label{tab:tinyimagenet}
\end{center}
\end{table*}

\subsubsection{Implementation Details}
The procedure of the experiments here closely follows the model compression benchmark \cite{tian2019crd}. The experiments include a large number of combinations between the teacher and student models. The teacher-student pairs include the models from the same architectural family but with different depth or width, and the models from different architectures that result in different sizes of features. The list of neural network architectures includes ResNet \cite{he2016deep}, WideResNet \cite{zagoruyko2016wide}, VGG \cite{simonyan2014very}, MobileNet \cite{howard2017mobilenets}, and ShuffleNet \cite{zhang2018shufflenet}. We use the feature map output of the last convolutional block for the features and the last linear layer's outputs for the logits. When the student's feature map is different from the teacher's, the transformation function $r$ resizes the feature maps spatially with PyTorch's pooling operation \cite{NEURIPS2019_9015}. Then, the student's number of channels is linearly projected by a 1x1-conv layer ($r$) to match the teacher's channel number. Lastly, the resulting feature map is flattened for $z$. For consistency, we include $r$ in all our variants that use features (but not logits), regardless of whether the feature maps have the same size or not.

For learning with the CIFAR100 dataset, the models have an initial learning rate of 0.05, decayed by 0.1 every 30 epochs after the first 150 epochs until it reaches 240. For MobileNetV2, ShuffleNetV1 and ShuffleNetV2, the initial learning rate is  0.01 as suggested by \cite{tian2019crd}. All the methods use SGD with a momentum of 0.9 and a batch size of 64. In short, we follow the benchmark settings \cite{RepDistiller}
and use the same teacher models provided to conduct all the experiments, ensuring a fair comparison between all methods.

\textbf{Hyper-parameter selection} could have a profound effect on most knowledge distillation-based model compression. We follow the benchmark protocol \cite{tian2019crd}, which selects the hyper-parameters based on only one teacher-student pair (we use resnet32x4/resnet8x4), then apply it to all other cases. Therefore, a method has to be robust to the hyper-parameters choice to perform well in all cases. 

We make a step further to align the hyperparameter $\lambda$ used in Features-SE and Weighted$_E$ Features-SE. This can be achieved by normalizing the $W(\Zt)$'s outputs to make its diagonal to have a mean of one (like the identity matrix $\mathbb{I}$) and unit variance. This normalization makes the features have an expected importance of 1 no matter how $W$ is computed, leaving the gradient-based weighting the only factor to affect the performance between the two cases. As a result, we can use the coefficient $\lambda=\lambda_F=3$ in all cases for features. Additionally, $\lambda=\lambda_L=15$ when we use logits.
Lastly, we add an extra setting by combining features and logits. It leads to the case of "B+C" in Tables \ref{tbl:cifar100_same} and \ref{tbl:cifar100_diff} with the custmoized $D_G$:
\begin{align} \label{eq:D_G_MC_FL}
D_{G-BC}^{MC}&=\lambda_L (\hat{l}_s-\hat{l}_t)^T (\hat{l}_s-\hat{l}_t)\nonumber\\
&+ \lambda_F (\hat{z}_s-\hat{z}_t)^T W_E(z_t) (\hat{z}_s-\hat{z}_t)
\end{align}
Note that $\hat{l}$ is the normalized logits and $\hat{z}$ is the normalized features. We use  $\lambda=1$ for $D_{G-BC}^{MC}$.

\subsubsection{MC Benchmark Results} \label{sec:MC_analysis}

First of all, we emphasize that our focus is on evaluating our general formulation and keeping their instantiations in the simplest form, revealing the intrinsic trend of KD with different knowledge sources. Our goal here is not beating state-of-the-art (\eg CRD), although our methods (B and C in Tables \ref{tbl:cifar100_same} and \ref{tbl:cifar100_diff}) achieve comparable or better performance.
In our comparison, Tables \ref{tbl:cifar100_same} and \ref{tbl:cifar100_diff} statistically agree with the arguments made at the beginning of the MC section:
First, the weighted features (B) performs better than the unweighted one (A) in 9 out of 11 cases.
Second, the logits (C) performs better than features (A and B) in 9 out of 11 cases.
This result suggests the rank of "logits $>$ weighted features $>$ plain features" in their KD efficiency. Besides, our variants (A, B, C) outperform most of the previous KD methods, showing that our methods are efficient in extracting out the knowledge, and making the observed ranks more representative.

\subsubsection{Experiment on ImageNet}

We use a larger and harder dataset for replicating the experiments of Table \ref{tbl:cifar100_same} with the standard ResNet. Table \ref{tab:tinyimagenet} uses the subset of ImageNet images \cite{TinyImageNet}
, showing the same trend of "logits $>$ weighted features $>$ plain features", and combing all (B+C) leads to the best result. Tiny-ImageNet has 200 classes sub-sampled from ImageNet. Each class has 500 training images and testing images with size 64x64. All the methods in Table \ref{tab:tinyimagenet} use the same hyperparameters as in Tables \ref{tbl:cifar100_same} and \ref{tbl:cifar100_diff}. The training configuration is similar to Tables \ref{tbl:cifar100_same}, except that the initial learning rate is 0.1, and is decayed by the factor of 0.1 at 50\% and 75\% of total (100) epochs. The weight-decay is 0.005. The models (ResNet-18/34/50) are the default models defined in the PyTorch. All students are trained from scratch (with random initialization). The teacher's weights were initialized with an ImageNet(full)-pretrained model in PyTorch model zoo, then is fine-tuned with Tiny-ImageNet. 

\begin{figure}
  \centering
  \includegraphics[clip, trim=0.5cm 0.65cm 0cm 0cm, width=0.49\textwidth]{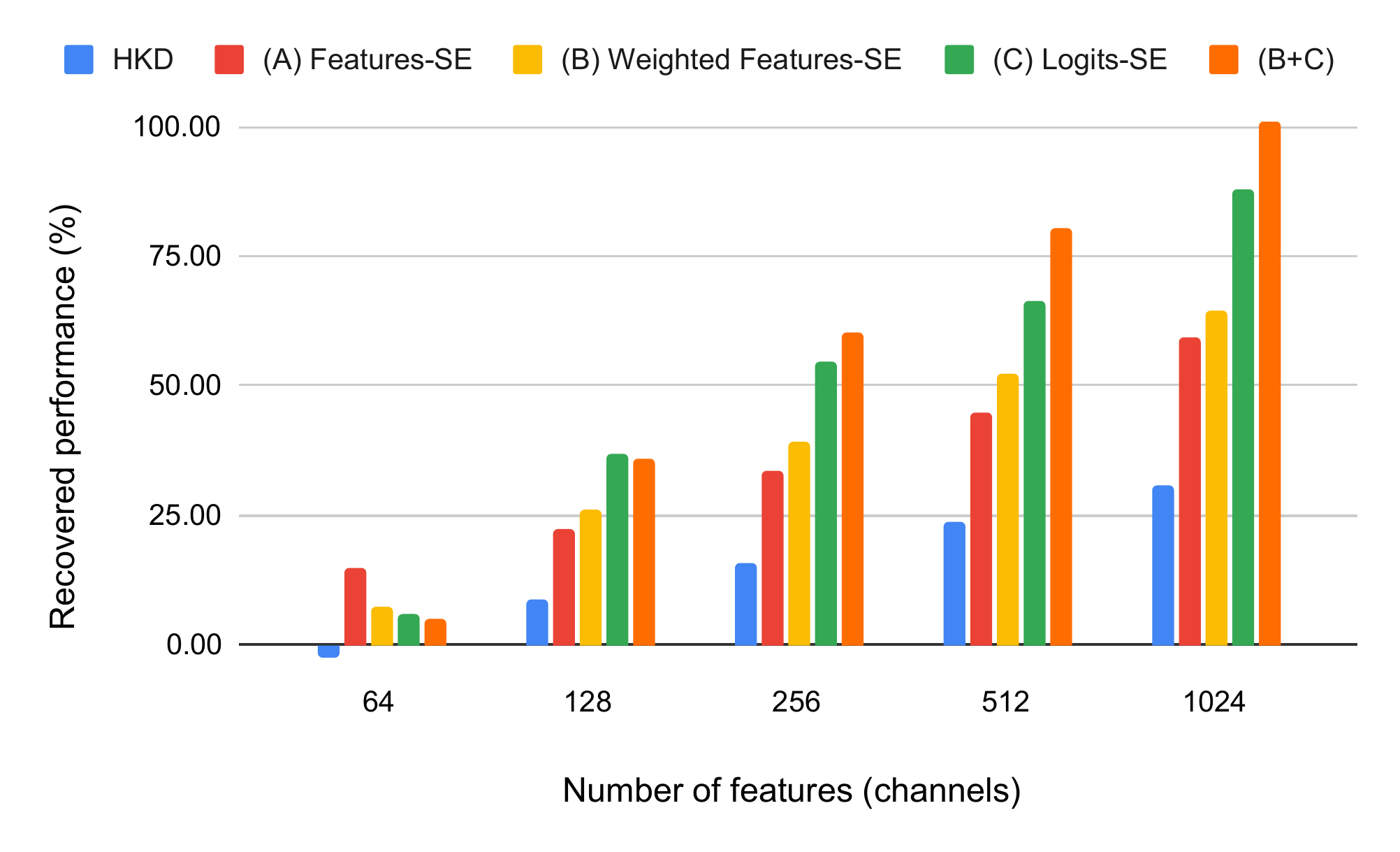} 
  \caption{
  The analysis of feature size versus sources of knowledge. 
  The performance gain due to model capacity has been subtracted.  This figure highlights the changes in the ranking of knowledge sources. The raw value of each bar is averaged with 5 repeats and is available in Supplementary.
  }\label{fig:width_effect}
\end{figure}

\subsubsection{Key factor analysis}\label{sec:key}

This section investigates the factor that affects the ranking of knowledge sources. The clue comes from Tables \ref{tbl:cifar100_same} and \ref{tbl:cifar100_diff}, in which Table \ref{tbl:cifar100_diff} has a larger difference on (B-A) and (C-B) than in Table \ref{tbl:cifar100_same}. One possible factor is that the students in Table \ref{tbl:cifar100_same} generally have a smaller feature size (\eg, resnet32: 64 channels; WRN-40-1: 64; WRN-40-2: 128) than the students in Table \ref{tbl:cifar100_diff} (\eg, MobileNetV2: 160; ShuffleNetV1: 960; ShuffleNetV2: 464).
To investigate whether the student's feature size has a substantial impact, we vary the feature size within the same model architecture. However, changing feature size inevitably changes the model's capacity, and a larger capacity helps the student more easily match the teacher. Thus, comparing the absolute accuracy can not make a conclusive analysis. 

We therefore make an additional contribution in addressing the above issue for the analysis, applying two strategies to minimize the capacity effect: (1) only the number of channels for the last convolutional block is changed, and (2) we propose a recovered performance ratio to measure the relative performance gain by subtracting the performance of a vanilla student (trained without a teacher). The vanilla student has its performance increased along with the capacity; therefore, subtracting its accuracy ($Acc^s_{vanilla}$) excludes the capacity effect. The recovered performance ratio (RPR) is computed by:
\begin{equation} \label{eq:RPR}
RPR=(Acc^s_{KD}-Acc^s_{vanilla})/(Acc^t_{vanilla}-Acc^s_{vanilla}). 
\end{equation}

Figure \ref{fig:width_effect} uses RPR to examine various student feature sizes. The teacher and student models are resnet32x4 and resnet8x4, correspondingly. The student model's last convolutional block has its width (number of channels) configured to be between 64 to 1024. The result confirms feature size's impact on the ranking:
\begin{itemize}
    \item A larger student feature (\eg $\geq256$) leads to a consistent ranking of "logits $>$ weighted features $>$ plain features". A smaller feature (\eg 64) breaks the trend.
\end{itemize}
We additionally subtract a stronger baseline (HKD \cite{hinton2014distilling}) by replacing \eqref{eq:RPR}'s $Acc^s_{vanilla}$ with $Acc^s_{HKD}$ in Supplementary.
Its trend is still the same as Figure \ref{fig:width_effect}, providing extra support for the observation.
The result suggests the design guideline:
\begin{itemize}
    \item Having a larger student feature size can benefit KD. Note that using larger features does not conflict with the goal of model compression. ShuffleNet \cite{zhang2018shufflenet} in Table \ref{tbl:cifar100_diff} is a positive case that has a relatively smaller model while still having a sufficient feature size.
\end{itemize}

\subsection{Incremental Learning (IL)} \label{sec:CL}

Incremental learning is a problem setting where KD is often applied. The setting has its model exposed to a sequence of tasks. These tasks have differences in either their input distribution, label distribution, or both. The model has no access to the training data of previous tasks when learning a new task. The shift of distributions among tasks introduces a significant interference to the learned parameters, largely undermining previous tasks' performance. This phenomenon is called catastrophic forgetting. A popular strategy is to regularize the model's parameters to mitigate the forgetting, reducing drift from its previously learned parameters. However, when the regularization is too strong, the model will not have sufficient plasticity to learn a new task well. Thus, there is a trade-off between minimizing forgetting and maximizing plasticity.
A good trade-off strategy keeps important knowledge while allowing the less important ones to be overwritten by the new tasks. The parameter-based regularization dominates this line of strategy. Previous works \cite{kirkpatrick2017overcoming,zenke2017continual,aljundi2017memory} select important parameters based on gradients and avoid those parameters from changing too much. It is a setting that complements model compression, providing an excellent opportunity to \textit{compare not only features, logits, and gradients, but also the model parameters for transferring the knowledge}.

\subsubsection{Applying our framework}

We consider the task-incremental learning \cite{Hsu18_EvalCL,van2019three} setting for our experiments. This setting has exclusive sets of classes in a sequence of classification tasks. The model learns each classification task sequentially with only access to the training data of the current task. During the learning curriculum, the model regularly adds an output head (as a linear layer) for a new classification task, while inheriting all the parts learned in the previous tasks. As a result, the model has multiple heads (one for each task), and it requires $D_G$ to sum over all previous tasks' output heads to regularize the model drifting. Specifically, $D_G$ is customized by:
\begin{equation}
D_{G-logits}^{IL}=\sum_j(l^s_{[j]}-l^t_{[j]})^T (l^s_{[j]}-l^t_{[j]})\label{eq:D_G_CL_logit}
\end{equation}
\begin{equation}
D_{G-features}^{IL}=\sum_j(\Zs-\Zt)^T W_{[j]}(\Zt) (\Zs-\Zt) \label{eq:D_G_CL_feat}
\end{equation}
\begin{equation}
W_{[j]}(\Zt)=diag((\frac{d}{dz} \frac{1}{k}\sum_{y=1}^k (l^t_{[j],y})^2)(\frac{d}{dz} \frac{1}{k}\sum_{y=1}^k (l^t_{[j],y})^2)^T)\label{eq:W_CL}
\end{equation}

The $l_{[j]}$ is the logits from the $j$th task. The regularization term sums over the tasks except the current task $T_{current}$  (\ie task index  $j=\{1..T_{current}-1\} $). Note that when $T_{current}=2$, everything here (equations \ref{eq:D_G_CL_logit} to \ref{eq:W_CL}) falls back to equations \ref{eq:D_G} and \ref{eq:W}. The only difference is that the current task's labels are out-of-scope for the previous (teacher) model. In other words, this is the case that $y^*$ is not valid for the teacher; therefore, \eqref{eq:W_CL} uses $\Ls_H$ to collect the gradients. In this section, our Logits-SE uses $D_{G-logits}^{IL}$, Weighted$_H$ Features-SE uses $D_{G-features}^{IL}$, and Features-SE has its $W_{[j]}(\Zt)=\mathbb{I}$. We additionally add three EWC variants for the comparison. SI \cite{zenke2017continual} accumulates the gradients along the optimization trajectory to replace the $F(\theta^t)$ in \eqref{eq:EWC}. MAS \cite{aljundi2017memory} uses $\Ls_H$ to compute the gradients for a weighting matrix similar to our $W$. The L2 sets its $F(\theta^t)=\mathbb{I}$ in \eqref{eq:EWC}.
All experiments here closely follow the implementation and evaluation protocol described in the popular benchmark
\cite{CLbenchmark}. 
More details are in Supplementary.

\begin{table}
\begin{small}
\begin{tabular}{lcccc}
\toprule
Method &Reg &S-CIFAR10 &S-CIFAR100 \\
\midrule
Upper-bound & - &97.6 $\pm$ 0.1 &84.3 $\pm$ 0.4 \\
Baseline &- &63.5 $\pm$ 1.7 &30.5 $\pm$ 0.6 \\
L2 &P &74.2 $\pm$ 0.6 &51.7 $\pm$ 1.3 \\
EWC (Kirkpatrick 2017)\tempcite{kirkpatrick2017overcoming} &P &84.4 $\pm$ 2.1 &61.1 $\pm$ 1.4 \\
SI (Zenke 2017)\tempcite{zenke2017continual} &P &79.1 $\pm$ 1.3 &64.8 $\pm$ 1.0 \\
MAS (Aljundi 2018)\tempcite{aljundi2017memory} &P &78.3 $\pm$ 0.7 &64.8 $\pm$ 0.8 \\
\midrule
Features-SE &F &77.4 $\pm$ 3.6 &70.5 $\pm$ 0.6 \\
Weighted$_H$ Features-SE &F &93.3 $\pm$ 0.5 &73.7 $\pm$ 0.6 \\
Logits-SE &L &\textbf{95.3} $\pm$ 0.2 &\textbf{78.3} $\pm$ 0.3 \\
\bottomrule
\end{tabular}
\end{small}
\caption{The averaged classification accuracy of task-incremental learning. The S-CIFAR10/100 datasets split their classes into 5 tasks for a model to learn sequentially. The regularization targets (Reg) are parameters (P), features (F), and logits (L). The upper-bound performance is from non-incremental learning (learn 5 tasks simultaneously).
}\label{tab:split_cifar}
\end{table}

\subsubsection{IL Benchmark Results}
Table \ref{tab:split_cifar} shows that the effectiveness of knowledge sources is ranked: logits (L) $>$ features (F) $>$ parameters (P). Although the number of classes imposes a very different difficulty to the problem (2 classes per task in S-CIFAR10 versus 20 in S-CIFAR100), the methods noted with "P" performs significantly worse than "F" and "L" on both datasets, suggesting that regularizing the outputs generally strikes a better balance between forgetting and plasticity. Furthermore, the comparison between Weighted$_H$ Features-SE versus Features-SE shows that having the squared error weighted by gradients is very helpful. Both above observations are consistent with the trends in model compression.

\section{Related Work}

We categorize KD methods by their knowledge sources and discuss the most related works. First, the features-based methods generally make a small student match a large teacher's features without selection. One exception is  \cite{heo2019comprehensive}, which selects useful features by using margin ReLU with a per-feature threshold. However, its heuristic nature is significantly different from our gradient-driven approach (\ie, Weighted Features-SE ). It is also worth noting that our Features-SE is closely related FitNet \cite{romero2014fitnets} and FT \cite{kim2018paraphrasing}. FitNet, FT, and our method align the dimension of the features between the teacher and student, then use squared error to match the features. However, FitNet adds a convolutional regressor (with non-linearity) for the student to do the matching, and trains the student with squared error loss and cross-entropy loss in two separate stages. In FT, it uses two small auto-encoders to transform the features from both the teacher and student. Therefore, both FitNet and FT have a more complicated design than our linear transformation function $r$ and our one-stage training procedure. Second, in previous gradient-based methods \cite{srinivas2018knowledge, zagoruyko2016paying}, they directly match the teacher's and student's Jacobian, which requires double backpropagation to optimize its loss function. In contrast, our weighted features-SE does not use Jacobian, avoiding the heavy overhead in optimization. Lastly, the logits-based methods \cite{hinton2014distilling, li2016learning} have been discussed in previous sections. Other logits-based strategies such as early stopping \cite{cho2019efficacy} and teacher assistant \cite{mirzadeh2020improved} are orthogonal approaches to our work and can be applied jointly.

\section{Conclusion}

We present a new perspective that can utilize different knowledge sources under a unified KD framework. This framework leads to a new KD method that prioritizes the distillation of important features based on gradients, and provides a new justification on how simple squared error approximates the classical KD criteria. We instantiate our framework based on the type of knowledge sources utilized, finding that logits is generally more efficient than features, while gradients can help the latter. Furthermore, our analysis points out that a student's feature size is impactful to the KD efficiency. We hope the new insights, new methods, and new findings will inspire more works in this field.

\bibliography{egbib}

\twocolumn[
  \begin{@twocolumnfalse}
    \newpage
    \null
    \vskip .375in
    \begin{center}
      {\LARGE \bf A Closer Look at Knowledge Distillation with Features, Logits, and Gradients\\-Supplementary Materials- \par}
      \vspace*{24pt}
      {
      \large
      \lineskip .5em
      \begin{tabular}[t]{c}
    Paper ID 2350 
      \end{tabular}
      \par
      }
      \vskip .5em
      \vspace*{12pt}
    \end{center}
  \end{@twocolumnfalse}
]
\setcounter{section}{0}
\setcounter{equation}{0}
\setcounter{figure}{0}
\setcounter{table}{0}
\setcounter{page}{1}
\makeatletter
\renewcommand\thesection{S\arabic{section}}
\renewcommand{\theequation}{S\arabic{equation}}
\renewcommand{\thefigure}{\Alph{figure}}
\renewcommand{\thetable}{\Alph{table}}







\section{Detailed Derivations of Equation \ref{eq:KLrawexpand}}
This section provides the detailed steps that are not included in the main text for deriving \eqref{eq:KLapprox} from \eqref{eq:KLrawexpand}.

(1) The first-order term in \eqref{eq:KLrawexpand} is zero:
\begin{align*}
-dz^T\sum_y \Pt_y \frac{d}{dz}\log \Pt_y &= -dz^T \sum_y \frac{d}{dz}\Pt_y\\
&= -dz^T (\frac{d}{dz} \sum_y \Pt_y)\\
&= 0
\end{align*}

(2) The second-order term in \eqref{eq:KLrawexpand} has a form of Fisher information matrix $F(\Zt)$ at its middle:
\begin{align*}
- & \frac{1}{2}dz^T (\sum_y\Pt_y\frac{d^2}{dz^2}\log\Pt_y) dz \\
&= -\frac{1}{2}dz^T\sum_y\Pt_y\frac{d}{dz}(\frac{1}{\Pt_y}\frac{d\Pt_y}{dz}) dz\\
&= -\frac{1}{2}dz^T\sum_y\Pt_y \left[ \frac{1}{\Pt_y}\frac{d^2\Pt_y}{dz^2}-(\frac{1}{\Pt_y}\frac{d\Pt_y}{dz})(\frac{1}{\Pt_y}\frac{d\Pt_y}{dz})^T \right] dz\\
&= -\frac{1}{2}dz^T \left[ \sum_y \frac{d^2\Pt_y}{dz^2} - \sum_y \Pt_y (\frac{1}{\Pt_y}\frac{d\Pt_y}{dz})(\frac{1}{\Pt_y}\frac{d\Pt_y}{dz})^T \right] dz\\
&= -\frac{1}{2}dz^T \left[ \frac{d^2}{dz^2} \sum_y \Pt_y - \sum_y \Pt_y (\frac{1}{\Pt_y}\frac{d\Pt_y}{dz})(\frac{1}{\Pt_y}\frac{d\Pt_y}{dz})^T \right] dz\\
&= -\frac{1}{2}dz^T \left[-\sum_y \Pt_y (\frac{1}{\Pt_y}\frac{d\Pt_y}{dz})(\frac{1}{\Pt_y}\frac{d\Pt_y}{dz})^T \right] dz\\
&= \frac{1}{2}dz^T \left[\sum_y \Pt_y (\frac{d}{dz} \log \Pt_y)(\frac{d}{dz} \log \Pt_y)^T \right] dz\\
&= \frac{1}{2}dz^T F(\Zt) dz
\end{align*}

\section{Additional analysis of Figure \ref{fig:width_effect}}

\begin{figure}
  \centering
  \includegraphics[clip, trim=0.5cm 0.5cm 0cm 0cm, width=0.47\textwidth]{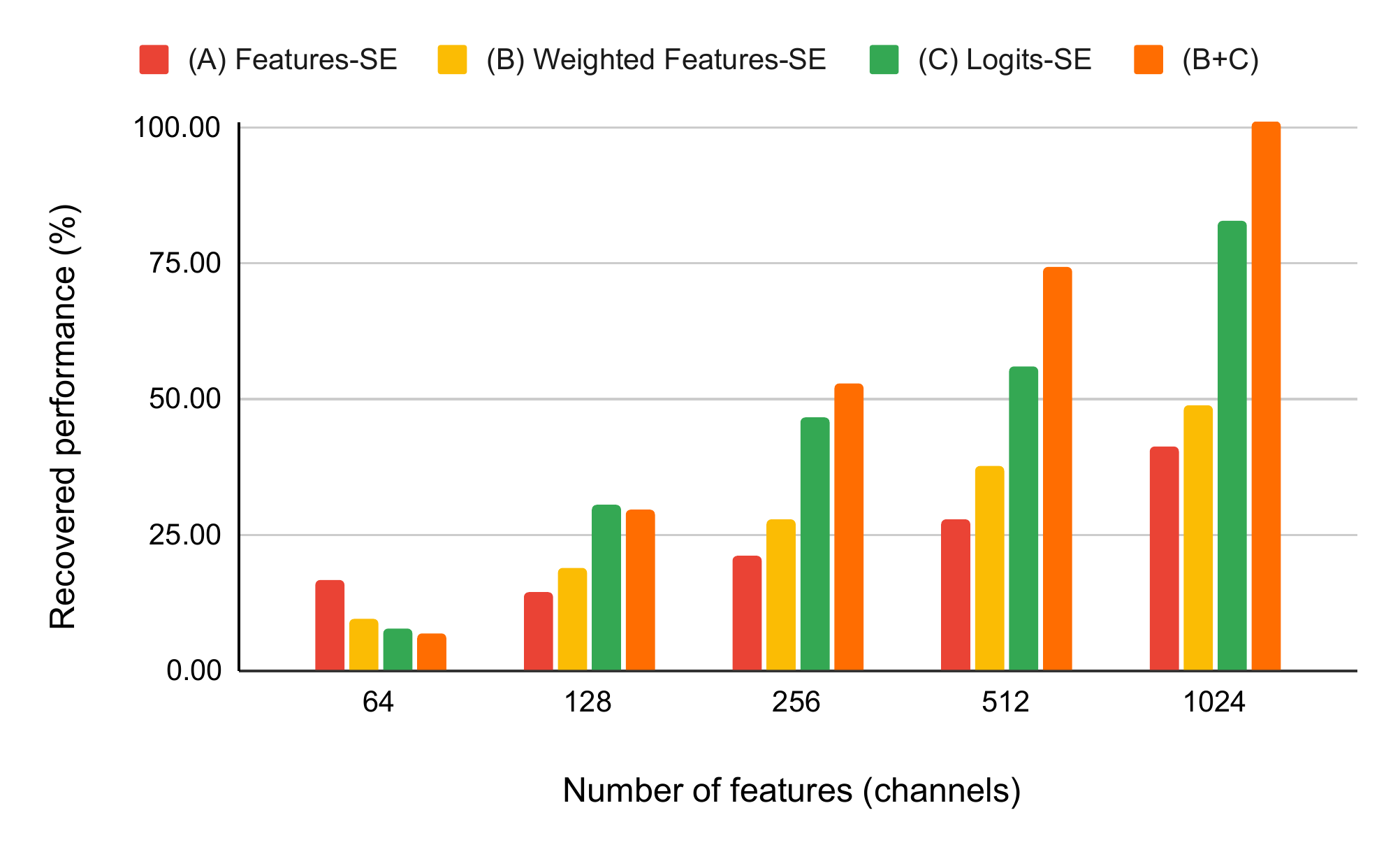} 
  \caption{The analysis of feature size versus sources of knowledge. The performance gain due to model capacity has been more aggressively removed by subtracting the performance of HKD ~\cite{hinton2014distilling} in the RPR (\eqref{eq:RPR}). The ranking of methods is still consistent with Figure \ref{fig:width_effect}.
  }\label{fig:width_effect_base_hkd}
\end{figure}

Figure \ref{fig:width_effect_base_hkd} has the same experiment as in Figure \ref{fig:width_effect}, but the performance gain due to model capacity has been more aggressively removed by subtracting the performance of HKD ~\cite{hinton2014distilling} in the RPR (\eqref{eq:RPR}). This figure has a trend similar to Figure \ref{fig:width_effect}, strengthening the argument that the student's feature size has a crucial impact on knowledge distillation. Table \ref{tab:width_raw} provides the raw accuracy used in both Figures \ref{fig:width_effect} and \ref{fig:width_effect_base_hkd}.

\begin{table*}
\begin{center}
\begin{small}
\begin{tabular}{cccccccc}\toprule
\#channel     &Baseline &HKD &(A) Features-SE &(B) Weighted Feature-SE &(C) Logits-SE &(B+C) \\\midrule
64   &68.12 &67.87 $\pm$ 0.18 &69.79 $\pm$ 0.36 &68.96 $\pm$ 0.47 &68.77 $\pm$ 0.14 &68.67 $\pm$ 0.49 \\
128  &70.55 &71.34 $\pm$ 0.43 &72.52 $\pm$ 0.25 &72.88 $\pm$ 0.12 &73.82 $\pm$ 0.28 &73.73 $\pm$ 0.24 \\
256  &72.50 &73.33 $\pm$ 0.25 &74.83 $\pm$ 0.15 &75.20 $\pm$ 0.15 &76.29 $\pm$ 0.16 &76.66 $\pm$ 0.22 \\
512  &73.62 &74.98 $\pm$ 0.17 &76.22 $\pm$ 0.10 &76.66 $\pm$ 0.13 &77.47 $\pm$ 0.25 &78.29 $\pm$ 0.20 \\
1024 &74.40 &75.94 $\pm$ 0.24 &77.38 $\pm$ 0.20 &77.64 $\pm$ 0.19 &78.82 $\pm$ 0.08 &79.47 $\pm$ 0.25 \\
\bottomrule
\end{tabular}
\end{small}
\caption{The raw accuracy (CIFAR100) used in Figures \ref{fig:width_effect} and \ref{fig:width_effect_base_hkd}. The baseline is the student model trained without knowledge distillation. }\label{tab:width_raw}
\end{center}
\end{table*}

\begin{table*}
\begin{center}
\begin{small}
\begin{tabular}{lccccc}
\toprule
Teacher model         & resnet32x4       & vgg13            & WRN-40-2         & WRN-40-2         & resnet110          \\
Student model         & resnet8x4        & vgg8             & WRN-16-2         & WRN-40-1         & resnet32           \\
\midrule
Teacher size (M)         & 7.43            & 9.46            & 2.26            & 2.26             & 1.74              \\
Student size (M)         & 1.23            & 3.96            & 0.70            & 0.57             & 0.47              \\
Compression rate         & 0.17            & 0.42            & 0.31            & 0.25             & 0.27              \\
\bottomrule						
\end{tabular}
\end{small}
\caption{
The size of models used in Table \ref{tbl:cifar100_same}
}
\label{tbl:cifar100_same_size}
\end{center}
\end{table*}

\begin{table*}[!htbp]
\begin{center}
\begin{small}
\begin{tabular}{lcccccc}
\toprule
Teacher model         & resnet32x4       & resnet32x4       & WRN-40-2         & ResNet50         & ResNet50         & vgg13         \\
Student model         & ShuffleNetV1     & ShuffleNetV2     & ShuffleNetV1     & vgg8             & MobileNetV2      & MobileNetV2   \\
\midrule
Teacher size (M)      & 7.43             & 7.43             & 2.26             & 23.71            & 23.71            & 9.46          \\
Student size (M)      & 0.95             & 1.36             & 0.95             & 3.96             & 0.81             & 0.81          \\
Compression rate      & 0.13             & 0.18             & 0.42             & 0.17             & 0.03             & 0.09          \\
\bottomrule						
\end{tabular}
\end{small}
\caption{
The size of models used in Table \ref{tbl:cifar100_diff}
}
\label{tbl:cifar100_diff_size}
\end{center}
\end{table*}

\section{Model Compression Size}

Tables \ref{tbl:cifar100_same_size} and \ref{tbl:cifar100_diff_size} provide the size of each model used in Tables \ref{tbl:cifar100_same} and \ref{tbl:cifar100_diff}.

\section{Implementation Details of IL Experiment \ref{sec:CL}} 
Our task-incremental learning experiment follows the implementation and evaluation protocol described in a popular continual learning benchmark~\cite{Hsu18_EvalCL}. The benchmark splits the image datasets CIFAR10 and CIFAR100 into 5 tasks; thus, each task has 2 and 20 classes, correspondingly. The evaluation is performed at the end of the learning curriculum, and the averaged classification accuracy of all tasks is reported with testing data. The regularization coefficient $\lambda$ of all methods (except the baseline and non-incremental learning) are selected by a grid search with 20\% of the dataset. We follow all the default training and testing configurations provided by the public benchmark \cite{CLbenchmark} to conduct the experiments. Lastly, In our methods, the features used in Weighted$_H$ Feature-SE and Feature-SE are the flattened outputs from the last convolutional block of WideResNet-28-2 \cite{zagoruyko2016wide}, which has the output dimension of 2048 (CxWxH=128x4x4). Note that we do not use the linear transformation function $r$ in Section \ref{sec:CL} since the teacher and student models always have the same feature dimensions. 

\section{Additional discussion: Model Compression versus Incremental Learning}
This section highlights three differences between Model Compression (MC) and Incremental Learning (IL) in their problem settings. First, MC has its teacher share the same training data with the student. In contrast, the student in task-incremental learning has no access to the teacher's training data and is exposed to a new set of classes in the new task.

Second, MC only applies the KD process once, while the task-IL repeats the KD process multiple times based on the length of the task sequence.

Lastly, the task-incremental learning setting obeys closely the assumptions made in Section \ref{sec:rethinking}, while MC relaxes those a little bit. Although MC's formulation (\eqref{eq:KD_compression}) has the same form as \eqref{eq:ourKD}, it deviates from the assumptions made in \eqref{eq:KLexpand} in three ways: (1) the $dz$ might not be small, (2) the $z$ is normalized and linearly transformed, and (3) the softmax-based classifier $f$ is not forced to be the same between the teacher and student.
Note that (2) and (3) happens when z is features but not logits, and the impact of (1) has been reduced by the normalization (\eqref{eq:norm_Z}).

\end{document}